\pdfoutput=1

\documentclass[11pt]{article}

\usepackage[final]{ACL2023}

\usepackage{times}
\usepackage{latexsym}

\usepackage[T1]{fontenc}
\usepackage[utf8]{inputenc}

\usepackage{microtype}
\usepackage{caption}
\usepackage{enumitem}
\usepackage{rotating}
\usepackage{microtype}
\usepackage{bm}
\usepackage{soul}
\usepackage{wrapfig}
\usepackage{balance}
\usepackage[varqu]{zi4}
\usepackage{units}
\usepackage{booktabs}
\usepackage{colortbl}
\usepackage{helvet}
\usepackage{printlen} %
\usepackage{placeins} %
\usepackage{afterpage} %
\usepackage{csquotes}
\usepackage{hyperref}
\usepackage{setspace}

\usepackage{amsmath}
\usepackage{mathtools}
\usepackage{bbm}

\definecolor{redOV}{RGB}{255, 235, 238}
\definecolor{redI}{RGB}{255, 205, 210}
\definecolor{redII}{RGB}{239, 154, 154}
\definecolor{redIII}{RGB}{229, 115, 115}
\definecolor{redIV}{RGB}{239, 83, 80}
\definecolor{redV}{RGB}{244, 67, 54}
\definecolor{redVI}{RGB}{229, 57, 53}
\definecolor{redVII}{RGB}{211, 47, 47}
\definecolor{redVIII}{RGB}{198, 40, 40}
\definecolor{redIX}{RGB}{183, 28, 28}
\definecolor{redAI}{RGB}{255, 138, 128}
\definecolor{redAII}{RGB}{255, 82, 82}
\definecolor{redAIV}{RGB}{255, 23, 68}
\definecolor{redAVII}{RGB}{213, 0, 0}

\definecolor{pinkOV}{RGB}{252, 228, 236}
\definecolor{pinkI}{RGB}{248, 187, 208}
\definecolor{pinkII}{RGB}{244, 143, 177}
\definecolor{pinkIII}{RGB}{240, 98, 146}
\definecolor{pinkIV}{RGB}{236, 64, 122}
\definecolor{pinkV}{RGB}{233, 30, 99}
\definecolor{pinkVI}{RGB}{216, 27, 96}
\definecolor{pinkVII}{RGB}{194, 24, 91}
\definecolor{pinkVIII}{RGB}{173, 20, 87}
\definecolor{pinkIX}{RGB}{136, 14, 79}
\definecolor{pinkAI}{RGB}{255, 128, 171}
\definecolor{pinkAII}{RGB}{255, 64, 129}
\definecolor{pinkAIV}{RGB}{245, 0, 87}
\definecolor{pinkAVII}{RGB}{197, 17, 98}

\definecolor{purpleOV}{RGB}{243, 229, 245}
\definecolor{purpleI}{RGB}{225, 190, 231}
\definecolor{purpleII}{RGB}{206, 147, 216}
\definecolor{purpleIII}{RGB}{186, 104, 200}
\definecolor{purpleIV}{RGB}{171, 71, 188}
\definecolor{purpleV}{RGB}{156, 39, 176}
\definecolor{purpleVI}{RGB}{142, 36, 170}
\definecolor{purpleVII}{RGB}{123, 31, 162}
\definecolor{purpleVIII}{RGB}{106, 27, 154}
\definecolor{purpleIX}{RGB}{74, 20, 140}
\definecolor{purpleAI}{RGB}{234, 128, 252}
\definecolor{purpleAII}{RGB}{224, 64, 251}
\definecolor{purpleAIV}{RGB}{213, 0, 249}
\definecolor{purpleAVII}{RGB}{170, 0, 255}

\definecolor{deeppurpleOV}{RGB}{237, 231, 246}
\definecolor{deeppurpleI}{RGB}{209, 196, 233}
\definecolor{deeppurpleII}{RGB}{179, 157, 219}
\definecolor{deeppurpleIII}{RGB}{149, 117, 205}
\definecolor{deeppurpleIV}{RGB}{126, 87, 194}
\definecolor{deeppurpleV}{RGB}{103, 58, 183}
\definecolor{deeppurpleVI}{RGB}{94, 53, 177}
\definecolor{deeppurpleVII}{RGB}{81, 45, 168}
\definecolor{deeppurpleVIII}{RGB}{69, 39, 160}
\definecolor{deeppurpleIX}{RGB}{49, 27, 146}
\definecolor{deeppurpleAI}{RGB}{179, 136, 255}
\definecolor{deeppurpleAII}{RGB}{124, 77, 255}
\definecolor{deeppurpleAIV}{RGB}{101, 31, 255}
\definecolor{deeppurpleAVII}{RGB}{98, 0, 234}

\definecolor{indigoOV}{RGB}{232, 234, 246}
\definecolor{indigoI}{RGB}{197, 202, 233}
\definecolor{indigoII}{RGB}{159, 168, 218}
\definecolor{indigoIII}{RGB}{121, 134, 203}
\definecolor{indigoIV}{RGB}{92, 107, 192}
\definecolor{indigoV}{RGB}{63, 81, 181}
\definecolor{indigoVI}{RGB}{57, 73, 171}
\definecolor{indigoVII}{RGB}{48, 63, 159}
\definecolor{indigoVIII}{RGB}{40, 53, 147}
\definecolor{indigoIX}{RGB}{26, 35, 126}
\definecolor{indigoAI}{RGB}{140, 158, 255}
\definecolor{indigoAII}{RGB}{83, 109, 254}
\definecolor{indigoAIV}{RGB}{61, 90, 254}
\definecolor{indigoAVII}{RGB}{48, 79, 254}

\definecolor{blueOV}{RGB}{227, 242, 253}
\definecolor{blueI}{RGB}{187, 222, 251}
\definecolor{blueII}{RGB}{144, 202, 249}
\definecolor{blueIII}{RGB}{100, 181, 246}
\definecolor{blueIV}{RGB}{66, 165, 245}
\definecolor{blueV}{RGB}{33, 150, 243}
\definecolor{blueVI}{RGB}{30, 136, 229}
\definecolor{blueVII}{RGB}{25, 118, 210}
\definecolor{blueVIII}{RGB}{21, 101, 192}
\definecolor{blueIX}{RGB}{13, 71, 161}
\definecolor{blueAI}{RGB}{130, 177, 255}
\definecolor{blueAII}{RGB}{68, 138, 255}
\definecolor{blueAIV}{RGB}{41, 121, 255}
\definecolor{blueAVII}{RGB}{41, 98, 255}

\definecolor{lightblueOV}{RGB}{225, 245, 254}
\definecolor{lightblueI}{RGB}{179, 229, 252}
\definecolor{lightblueII}{RGB}{129, 212, 250}
\definecolor{lightblueIII}{RGB}{79, 195, 247}
\definecolor{lightblueIV}{RGB}{41, 182, 246}
\definecolor{lightblueV}{RGB}{3, 169, 244}
\definecolor{lightblueVI}{RGB}{3, 155, 229}
\definecolor{lightblueVII}{RGB}{2, 136, 209}
\definecolor{lightblueVIII}{RGB}{2, 119, 189}
\definecolor{lightblueIX}{RGB}{1, 87, 155}
\definecolor{lightblueAI}{RGB}{128, 216, 255}
\definecolor{lightblueAII}{RGB}{64, 196, 255}
\definecolor{lightblueAIV}{RGB}{0, 176, 255}
\definecolor{lightblueAVII}{RGB}{0, 145, 234}

\definecolor{cyanOV}{RGB}{224, 247, 250}
\definecolor{cyanI}{RGB}{178, 235, 242}
\definecolor{cyanII}{RGB}{128, 222, 234}
\definecolor{cyanIII}{RGB}{77, 208, 225}
\definecolor{cyanIV}{RGB}{38, 198, 218}
\definecolor{cyanV}{RGB}{0, 188, 212}
\definecolor{cyanVI}{RGB}{0, 172, 193}
\definecolor{cyanVII}{RGB}{0, 151, 167}
\definecolor{cyanVIII}{RGB}{0, 131, 143}
\definecolor{cyanIX}{RGB}{0, 96, 100}
\definecolor{cyanAI}{RGB}{132, 255, 255}
\definecolor{cyanAII}{RGB}{24, 255, 255}
\definecolor{cyanAIV}{RGB}{0, 229, 255}
\definecolor{cyanAVII}{RGB}{0, 184, 212}

\definecolor{tealOV}{RGB}{224, 242, 241}
\definecolor{tealI}{RGB}{178, 223, 219}
\definecolor{tealII}{RGB}{128, 203, 196}
\definecolor{tealIII}{RGB}{77, 182, 172}
\definecolor{tealIV}{RGB}{38, 166, 154}
\definecolor{tealV}{RGB}{0, 150, 136}
\definecolor{tealVI}{RGB}{0, 137, 123}
\definecolor{tealVII}{RGB}{0, 121, 107}
\definecolor{tealVIII}{RGB}{0, 105, 92}
\definecolor{tealIX}{RGB}{0, 77, 64}
\definecolor{tealAI}{RGB}{167, 255, 235}
\definecolor{tealAII}{RGB}{100, 255, 218}
\definecolor{tealAIV}{RGB}{29, 233, 182}
\definecolor{tealAVII}{RGB}{0, 191, 165}

\definecolor{greenOV}{RGB}{232, 245, 233}
\definecolor{greenI}{RGB}{200, 230, 201}
\definecolor{greenII}{RGB}{165, 214, 167}
\definecolor{greenIII}{RGB}{129, 199, 132}
\definecolor{greenIV}{RGB}{102, 187, 106}
\definecolor{greenV}{RGB}{76, 175, 80}
\definecolor{greenVI}{RGB}{67, 160, 71}
\definecolor{greenVII}{RGB}{56, 142, 60}
\definecolor{greenVIII}{RGB}{46, 125, 50}
\definecolor{greenIX}{RGB}{27, 94, 32}
\definecolor{greenAI}{RGB}{185, 246, 202}
\definecolor{greenAII}{RGB}{105, 240, 174}
\definecolor{greenAIV}{RGB}{0, 230, 118}
\definecolor{greenAVII}{RGB}{0, 200, 83}

\definecolor{lightgreenOV}{RGB}{241, 248, 233}
\definecolor{lightgreenI}{RGB}{220, 237, 200}
\definecolor{lightgreenII}{RGB}{197, 225, 165}
\definecolor{lightgreenIII}{RGB}{174, 213, 129}
\definecolor{lightgreenIV}{RGB}{156, 204, 101}
\definecolor{lightgreenV}{RGB}{139, 195, 74}
\definecolor{lightgreenVI}{RGB}{124, 179, 66}
\definecolor{lightgreenVII}{RGB}{104, 159, 56}
\definecolor{lightgreenVIII}{RGB}{85, 139, 47}
\definecolor{lightgreenIX}{RGB}{51, 105, 30}
\definecolor{lightgreenAI}{RGB}{204, 255, 144}
\definecolor{lightgreenAII}{RGB}{178, 255, 89}
\definecolor{lightgreenAIV}{RGB}{118, 255, 3}
\definecolor{lightgreenAVII}{RGB}{100, 221, 23}

\definecolor{limeOV}{RGB}{249, 251, 231}
\definecolor{limeI}{RGB}{240, 244, 195}
\definecolor{limeII}{RGB}{230, 238, 156}
\definecolor{limeIII}{RGB}{220, 231, 117}
\definecolor{limeIV}{RGB}{212, 225, 87}
\definecolor{limeV}{RGB}{205, 220, 57}
\definecolor{limeVI}{RGB}{192, 202, 51}
\definecolor{limeVII}{RGB}{175, 180, 43}
\definecolor{limeVIII}{RGB}{158, 157, 36}
\definecolor{limeIX}{RGB}{130, 119, 23}
\definecolor{limeAI}{RGB}{244, 255, 129}
\definecolor{limeAII}{RGB}{238, 255, 65}
\definecolor{limeAIV}{RGB}{198, 255, 0}
\definecolor{limeAVII}{RGB}{174, 234, 0}

\definecolor{yellowOV}{RGB}{255, 253, 231}
\definecolor{yellowI}{RGB}{255, 249, 196}
\definecolor{yellowII}{RGB}{255, 245, 157}
\definecolor{yellowIII}{RGB}{255, 241, 118}
\definecolor{yellowIV}{RGB}{255, 238, 88}
\definecolor{yellowV}{RGB}{255, 235, 59}
\definecolor{yellowVI}{RGB}{253, 216, 53}
\definecolor{yellowVII}{RGB}{251, 192, 45}
\definecolor{yellowVIII}{RGB}{249, 168, 37}
\definecolor{yellowIX}{RGB}{245, 127, 23}
\definecolor{yellowAI}{RGB}{255, 255, 141}
\definecolor{yellowAII}{RGB}{255, 255, 0}
\definecolor{yellowAIV}{RGB}{255, 234, 0}
\definecolor{yellowAVII}{RGB}{255, 214, 0}

\definecolor{amberOV}{RGB}{255, 248, 225}
\definecolor{amberI}{RGB}{255, 236, 179}
\definecolor{amberII}{RGB}{255, 224, 130}
\definecolor{amberIII}{RGB}{255, 213, 79}
\definecolor{amberIV}{RGB}{255, 202, 40}
\definecolor{amberV}{RGB}{255, 193, 7}
\definecolor{amberVI}{RGB}{255, 179, 0}
\definecolor{amberVII}{RGB}{255, 160, 0}
\definecolor{amberVIII}{RGB}{255, 143, 0}
\definecolor{amberIX}{RGB}{255, 111, 0}
\definecolor{amberAI}{RGB}{255, 229, 127}
\definecolor{amberAII}{RGB}{255, 215, 64}
\definecolor{amberAIV}{RGB}{255, 196, 0}
\definecolor{amberAVII}{RGB}{255, 171, 0}

\definecolor{orangeOV}{RGB}{255, 243, 224}
\definecolor{orangeI}{RGB}{255, 224, 178}
\definecolor{orangeII}{RGB}{255, 204, 128}
\definecolor{orangeIII}{RGB}{255, 183, 77}
\definecolor{orangeIV}{RGB}{255, 167, 38}
\definecolor{orangeV}{RGB}{255, 152, 0}
\definecolor{orangeVI}{RGB}{251, 140, 0}
\definecolor{orangeVII}{RGB}{245, 124, 0}
\definecolor{orangeVIII}{RGB}{239, 108, 0}
\definecolor{orangeIX}{RGB}{230, 81, 0}
\definecolor{orangeAI}{RGB}{255, 209, 128}
\definecolor{orangeAII}{RGB}{255, 171, 64}
\definecolor{orangeAIV}{RGB}{255, 145, 0}
\definecolor{orangeAVII}{RGB}{255, 109, 0}

\definecolor{deeporangeOV}{RGB}{251, 233, 231}
\definecolor{deeporangeI}{RGB}{255, 204, 188}
\definecolor{deeporangeII}{RGB}{255, 171, 145}
\definecolor{deeporangeIII}{RGB}{255, 138, 101}
\definecolor{deeporangeIV}{RGB}{255, 112, 67}
\definecolor{deeporangeV}{RGB}{255, 87, 34}
\definecolor{deeporangeVI}{RGB}{244, 81, 30}
\definecolor{deeporangeVII}{RGB}{230, 74, 25}
\definecolor{deeporangeVIII}{RGB}{216, 67, 21}
\definecolor{deeporangeIX}{RGB}{191, 54, 12}
\definecolor{deeporangeAI}{RGB}{255, 158, 128}
\definecolor{deeporangeAII}{RGB}{255, 110, 64}
\definecolor{deeporangeAIV}{RGB}{255, 61, 0}
\definecolor{deeporangeAVII}{RGB}{221, 44, 0}

\definecolor{brownOV}{RGB}{239, 235, 233}
\definecolor{brownI}{RGB}{215, 204, 200}
\definecolor{brownII}{RGB}{188, 170, 164}
\definecolor{brownIII}{RGB}{161, 136, 127}
\definecolor{brownIV}{RGB}{141, 110, 99}
\definecolor{brownV}{RGB}{121, 85, 72}
\definecolor{brownVI}{RGB}{109, 76, 65}
\definecolor{brownVII}{RGB}{93, 64, 55}
\definecolor{brownVIII}{RGB}{78, 52, 46}
\definecolor{brownIX}{RGB}{62, 39, 35}

\definecolor{grayOV}{RGB}{250, 250, 250}
\definecolor{grayI}{RGB}{245, 245, 245}
\definecolor{grayII}{RGB}{238, 238, 238}
\definecolor{grayIII}{RGB}{224, 224, 224}
\definecolor{grayIV}{RGB}{189, 189, 189}
\definecolor{grayV}{RGB}{158, 158, 158}
\definecolor{grayVI}{RGB}{117, 117, 117}
\definecolor{grayVII}{RGB}{97, 97, 97}
\definecolor{grayVIII}{RGB}{66, 66, 66}
\definecolor{grayIX}{RGB}{33, 33, 33}

\definecolor{bluegrayOV}{RGB}{236, 239, 241}
\definecolor{bluegrayI}{RGB}{207, 216, 220}
\definecolor{bluegrayII}{RGB}{176, 190, 197}
\definecolor{bluegrayIII}{RGB}{144, 164, 174}
\definecolor{bluegrayIV}{RGB}{120, 144, 156}
\definecolor{bluegrayV}{RGB}{96, 125, 139}
\definecolor{bluegrayVI}{RGB}{84, 110, 122}
\definecolor{bluegrayVII}{RGB}{69, 90, 100}
\definecolor{bluegrayVIII}{RGB}{55, 71, 79}
\definecolor{bluegrayIX}{RGB}{38, 50, 56}
\definecolor{bluegrayX}{RGB}{17, 23, 26}

\definecolor{myACMBlue}{cmyk}{1,0.1,0,0.1}
\definecolor{myACMYellow}{cmyk}{0,0.16,1,0}
\definecolor{myACMOrange}{cmyk}{0,0.42,1,0.01}
\definecolor{myACMRed}{cmyk}{0,0.90,0.86,0}
\definecolor{myACMLightBlue}{cmyk}{0.49,0.01,0,0}
\definecolor{myACMGreen}{cmyk}{0.20,0,1,0.19}
\definecolor{myACMPurple}{cmyk}{0.55,1,0,0.15}
\definecolor{myACMDarkBlue}{cmyk}{1,0.58,0,0.21}

\definecolor{darkblue}{rgb}{0, 0, 0.5}
\colorlet{mylinkcolor}{darkblue}

\hypersetup{
  colorlinks,
  linkcolor=mylinkcolor,
  citecolor=mylinkcolor,
  urlcolor=mylinkcolor,
  filecolor=mylinkcolor
}
\newcommand{\link}[1]{{\href{#1}{\color{mylinkcolor}\textbf{\texttt{#1}}}}}
\newcommand{\linkhref}[2]{{\href{#1}{\color{mylinkcolor}{#2}}}}

\newcommand{\figpart}[1]{\textcolor{mylinkcolor}{#1}}

\newcommand{\tool}{{\textsc{WizMap}}}

\newcommand{\topheaderspace}{{\vspace{-1pt}}}
\newcommand{\botheaderspace}{{\vspace{-1pt}}}
\newcommand{\paraspace}{{\vspace{0pt}}}

\newcommand{\mapview}{\textit{Map View}}
\newcommand{\serachpanel}{\textit{Search Panel}}
\newcommand{\controlpanel}{\textit{Control Panel}}

\newcommand*{\vcenteredhbox}[1]{\begingroup\setbox0=\hbox{#1}\parbox{\wd0}{\box0}\endgroup}
\newcommand{\inlinefig}[2]{\vcenteredhbox{\includegraphics[height=#1pt]{figures/#2}}}

\title{\tool{}: Scalable Interactive Visualization for Exploring Large Machine Learning Embeddings}

\author{
  Zijie J. Wang \\
  Georgia Tech \\
  \texttt{jayw@gatech.edu} \\\And
  Fred Hohman \\
  Apple \\
  \texttt{fredhohman@apple.com} \\\And
  Duen Horng Chau \\
  Georgia Tech \\
  \texttt{polo@gatech.edu} \\
}

\makeatletter
\let\@oldmaketitle\@maketitle%
\renewcommand{\@maketitle}{\@oldmaketitle%
  \vspace{-27pt}
  \includegraphics[width=\linewidth]{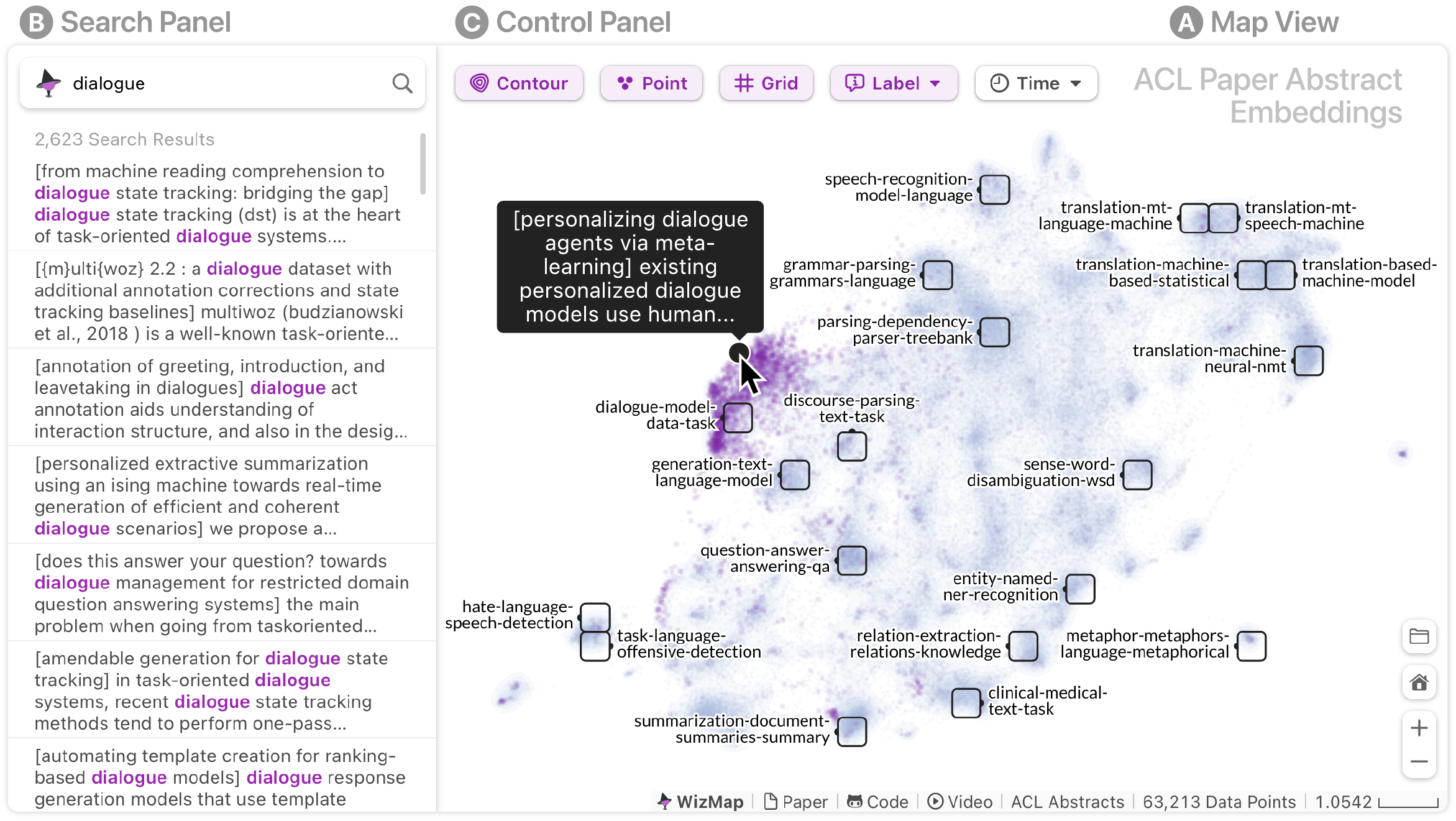}
  \vspace{-18pt}
  \captionof{figure}{
    \inlinefig{11}{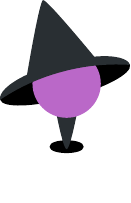}\hspace{2pt}\tool{} empowers machine learning researchers and domain experts to easily explore and interpret \textit{millions} of embedding vectors across different levels of granularity.
    Consider the task of investigating the embeddings of all 63k natural language processing paper abstracts indexed in ACL Anthology from 1980 to 2022.
    \textbf{(A) The Map View} tightly integrates a contour layer, a scatter plot, and automatically-generated multi-resolution embedding summaries to help users navigate through the large embedding space.
    \textbf{(B) The Search Panel} enables users to rapidly test their hypotheses through fast full-text embedding search.
    \textbf{(C) The Control Panel} allows users to customize embedding visualizations, compare multiple embedding groups, and observe how embeddings evolve over time.
  }
  \label{fig:crown}
  \vspace{17pt}
 }
\makeatother

\begin{document}
\maketitle

\begin{abstract}
  Machine learning models often learn latent embedding representations that capture the domain semantics of their training data.
  These embedding representations are valuable for interpreting trained models, building new models, and analyzing new datasets.
  However, interpreting and using embeddings can be challenging due to their opaqueness, high dimensionality, and the large size of modern datasets.
  To tackle these challenges, we present \tool{}, an interactive visualization tool to help researchers and practitioners easily explore large embeddings.
  With a novel multi-resolution embedding summarization method and a familiar map-like interaction design, \tool{} enables users to navigate and interpret embedding spaces with ease.
  Leveraging modern web technologies such as WebGL and Web Workers, \tool{} scales to millions of embedding points directly in users' web browsers and computational notebooks without the need for dedicated backend servers.
  \tool{} is open-source and available at the following public demo link: \link{https://poloclub.github.io/wizmap}.
\end{abstract}

\section{Introduction}
\botheaderspace{}

Modern machine learning (ML) models learn high-dimensional embedding representations to capture the domain semantics and relationships in the training data~\cite{raghuTransfusionUnderstandingTransfer2019}.
ML researchers and domain experts are increasingly using expressive embedding representations to interpret trained models~\cite{parkNeuroCartographyScalableAutomatic2022}, develop models for new domains~\cite{leeCleanNetTransferLearning2018} and modalities~\cite{ben-younesBLOCKBilinearSuperdiagonal2019}, as well as analyze and synthesize new datasets~\cite{kernGainingInsightsSocial2016}.
However, it can be difficult to interpret and use embeddings in practice, as these high-dimensional representations are often opaque, complex, and can contain unpredictable structures~\cite{bolukbasiManComputerProgrammer2016}.
Furthermore, analysts face scalability challenges as large datasets can require them to study millions of embeddings holistically~\cite{tangVisualizingLargescaleHighdimensional2016}.

To tackle these challenges, researchers have proposed several interactive visualization tools to help users explore embedding spaces~\cite[e.g.,][]{smilkovEmbeddingProjectorInteractive2016,liuLatentSpaceCartography2019}.
These tools often visualize embeddings in a low-dimensional scatter plot where users can browse, filter, and compare embedding points.
However, for large datasets, it is taxing or even implausible to inspect embedded data point by point to make sense of the \textit{global structure} of an embedding space.
Alternatively, recent research explores using contour plots to summarize embeddings~\cite{sevastjanovaVisualComparisonLanguage2022,robertsonAnglerHelpingMachine2023}.
Although contour abstractions enable users to obtain an overview of the embedding space and compare multiple embeddings through superposition, a user study reveals that contour plots restrict users' exploration of an embedding's \textit{local structures}, where users would prefer to have more visual context~\cite{robertsonAnglerHelpingMachine2023}.
To bridge this critical gap between two visualization approaches and provide users with a holistic view, we design and develop \tool{}~(\autoref{fig:crown}).
Our work makes the following \textbf{major contributions:}

\begin{figure}[tb]
  \includegraphics[width=\linewidth]{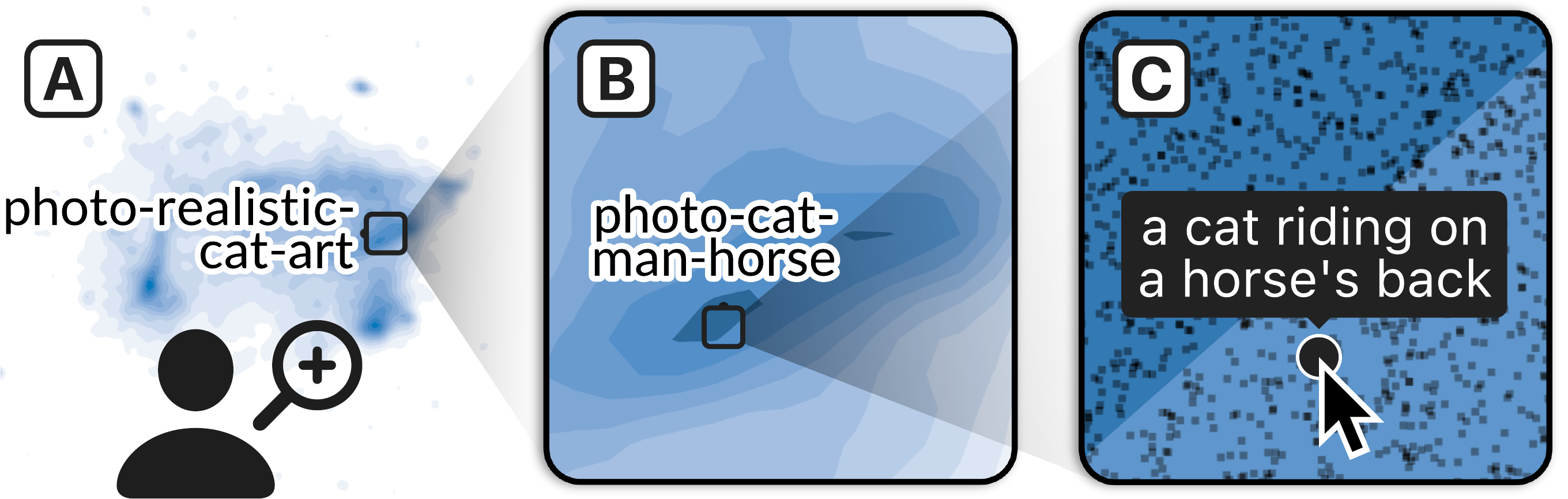}
  \caption{
    \tool{} enables users to explore embeddings at different levels of detail.
    \textbf{(A)} The contour plot with automatically-generated embedding summaries provides an overview.
    \textbf{(B)} Embedding summaries adjust in resolution as users zoom in.
    \textbf{(C)} The scatter plot enables the investigation of individual embeddings.
  }
  \label{fig:transition}
\end{figure}

\begin{itemize}[topsep=2pt, itemsep=0mm, parsep=3pt, leftmargin=9pt]
  \item \textbf{\tool{}, a scalable interactive visualization tool} that empowers ML researchers and domain experts to explore and interpret embeddings with \textit{millions} of points.
  Our tool employs a familiar map-like interaction design and fluidly presents adaptive visual summaries of embeddings across different levels of granularity~(\autoref{fig:transition}, \autoref{sec:interface}).

  \item \textbf{Novel and efficient method to generate multi-resolution embedding summaries.}
  To automatically summarize embedding neighborhoods with different degrees of granularity, we construct a quadtree~\cite{finkelQuadTreesData1974} from embedding points and extract keywords (text data) or exemplar points (other data types) from tree nodes with efficient branch aggregation~(\autoref{sec:method}).

  \item \textbf{An open-source\footnote{\tool{} code: \link{https://github.com/poloclub/wizmap}} and web-based implementation} that lowers the barrier to interpreting and using embeddings.
  We develop \tool{} with modern web technologies such as WebGL and Web Workers so that anyone can access the tool directly in both their web browsers and computational notebooks without a need for dedicated backend servers~(\autoref{sec:interface:implementation}).
  For a demo video of \tool{}, visit \link{https://youtu.be/8fJG87QVceQ}.
\end{itemize}

\setlength{\abovecaptionskip}{6pt}
\setlength{\belowcaptionskip}{-10pt}
\setlength{\abovecaptionskip}{10pt}
\setlength{\belowcaptionskip}{0pt} %
\section{Background and Related Work}
\botheaderspace{}

Researchers can extract a data point's embeddings by collecting its corresponding layer activations in neural networks trained for specific tasks such as classification and generation~\cite{raghuTransfusionUnderstandingTransfer2019}.
Additionally, researchers have developed task-agnostic models, such as word2vec~\cite{mikolovEfficientEstimationWord2013}, ELMo~\cite{petersDeepContextualizedWord2018}, and CLIP~\cite{radfordLearningTransferableVisual2021} that generate transferable embeddings directly.
These embeddings have been shown to outperform task-specific, state-of-the-art models in downstream tasks~\cite{radfordLearningTransferableVisual2021, dwibediLittleHelpMy2021}.

\subsection{Dimensionality Reduction}
\botheaderspace{}

Embeddings are often high-dimensional, such as \texttt{300}-dimensions for word2vec, or \texttt{768}-dimensions for CLIP and BERT Base~\cite{devlinBERTPretrainingDeep2018}.
Therefore, to make these embeddings easier to visualize, researchers often apply dimensionality reduction techniques to project them into 2D or 3D space.
Some popular dimensionality reduction techniques include UMAP~\cite{mcinnesUMAPUniformManifold2020}, t-SNE~\cite{vandermaatenVisualizingDataUsing2008}, and PCA~\cite{pearsonLinesPlanesClosest1901}.
Each of these techniques has its own strengths and weaknesses in terms of how well it preserves the embeddings' global structure, its stochasticity, interpretability, and scalability.
Despite these differences, all dimensionality reduction techniques produce data in the same structure.
This means users can choose any technique and visualize the projected embeddings with \tool{}.

\subsection{Interactive Embedding Visualization}
\botheaderspace{}

Researchers have introduced interactive visualization tools to help users explore embeddings~\cite[e.g.,][]{liuVisualExplorationSemantic2018, liEmbeddingVisVisualAnalytics2018, arendtParallelEmbeddingsVisualization2020}.
For example, Embedding Projector~\cite{smilkovEmbeddingProjectorInteractive2016} allows users to zoom, rotate, and pan 2D or 3D projected embeddings to explore and inspect data point features.
Similarly, Deepscatter~\cite{schmidtDeepscatterZoomableAnimated2021} and regl-scatterplot~\cite{lekschasReglScatterplotScalableInteractiveJavaScriptbased2023} empowers users to explore billion-scale 2D embeddings in their browsers.
Latent Space Cartography~\cite{liuLatentSpaceCartography2019} helps users find and refine meaningful semantic dimensions within the embedding space.
In addition, researchers have designed visualizations to aid users in comparing embeddings, such as embComp~\cite{heimerlEmbCompVisualInteractive2022} visualizing local and global similarities between two embeddings, Emblaze~\cite{sivaramanEmblazeIlluminatingMachine2022} tracing the changes in the position of data points across two embeddings, and Embedding Comparator~\cite{boggustEmbeddingComparatorVisualizing2022} highlighting the neighborhoods around points that change the most across embeddings.
In contrast, \tool{} aims to help users navigate and interpret both the global and local structures of large embedding spaces by offering visual contexts at varying levels of granularity.
\section{Multi-scale Embedding Summarization}
\label{sec:method}
\botheaderspace{}

Researchers have highlighted users' desire for embedding visualizations to provide visual contexts and embedding summaries to facilitate exploration of various regions within the embedding space~\cite{robertsonAnglerHelpingMachine2023}.
However, generating embedding summaries is challenging for two reasons.
First, efficiently summarizing millions of data points in larger datasets can be a formidable task.
Second, selecting the embedding regions to summarize is difficult, as users possess varying interests in regions of different sizes and levels of granularity.
To tackle this challenge, we propose a novel method to automatically generate multi-resolution embedding summaries at scale.

\paragraph{Multi-resolution Quadtree Aggregation.}
First, we apply a dimensionality reduction technique such as UMAP to project high-dimensional embedding vectors into 2D points.
From these points, we construct a quadtree~\cite{finkelQuadTreesData1974}, a tree data structure that recursively partitions a 2D space into four equally-sized squares, each represented as a node.
Each data point exists in a unique leaf node.
To summarize embeddings across different levels of granularity, we traverse the tree bottom up.
In each iteration, we first extract summaries of embeddings in each leaf node, and then merge the leaf nodes at the lowest level with their parent node.
This process continues recursively, with larger and larger leaf nodes being formed until the entire tree is merged into a single node at the root.
Finally, we map pre-computed embedding summaries to a suitable granularity level and dynamically show them as users zoom in or out in \tool{}~(\autoref{sec:interface:map:label}).

\setlength{\abovecaptionskip}{6pt}
\setlength{\belowcaptionskip}{-10pt}
\begin{figure}[tb]
  \includegraphics[width=\linewidth]{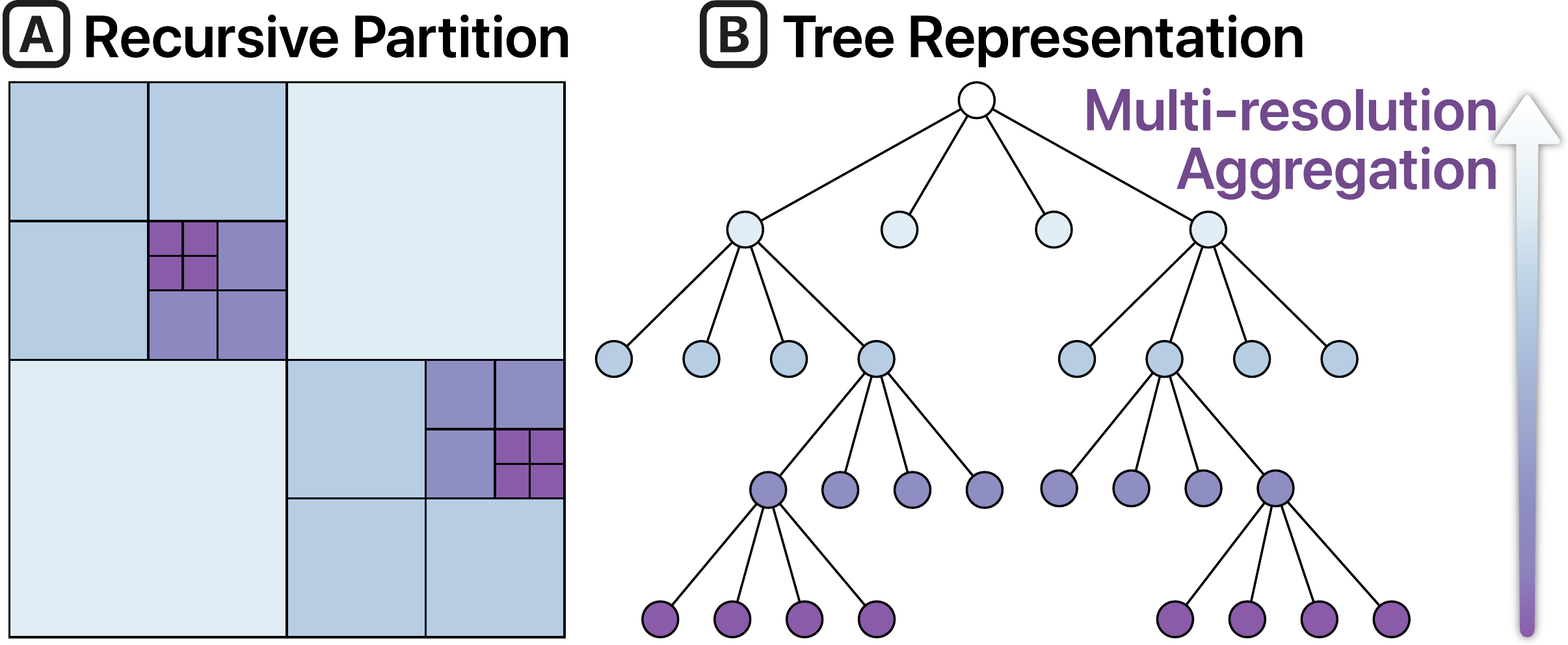}
  \caption{
    \textbf{(A)} A quadtree recursively partitions a 2D space into four equally-sized squares, \textbf{(B)} and each square is represented as a tree node.
    \tool{} efficiently aggregates information from the leaves to the root, summarizing embeddings at different levels of granularity.
  }
  \label{fig:quadtree}
\end{figure}
\setlength{\abovecaptionskip}{10pt}
\setlength{\belowcaptionskip}{0pt}

\paragraph{Scalable Leaf-level Summarization.}
When performing quadtree aggregation, researchers have the flexibility to choose any suitable method for summarizing embedding from leaf nodes.
For text embeddings, we propose t-TF-IDF (tile-based TF-IDF) that adapts TF-IDF (term frequency-inverse document frequency) to extract keywords from leaf nodes~\cite{sparckjonesStatisticalInterpretationTerm1972}.
Our approach is similar to c-TF-IDF (classed-based TF-IDF) that combines documents in a cluster into a meta-document before computing TF-IDF scores~\cite{grootendorstBERTopicNeuralTopic2022}.
Here, we merge all documents in each leaf node (i.e., a tile in the quadtree partition) as a meta-document and compute TF-IDF scores across all leaf nodes.
Finally, we extract keywords with the highest t-TF-IDF scores to summarize embeddings in a leaf node.
This approach is scalable and complementary to quadtree aggregation.
Because our document merging is hierarchical, we only construct the n-gram count matrix once and update it in each aggregation iteration with just one matrix multiplication.
Summarizing 1.8 million text embeddings across three granularity levels takes only about 55 seconds on a MacBook Pro.
For non-text data, we summarize embeddings by finding points closest to the embedding centroid in a leaf node.
\topheaderspace{}
\section{User Interface}
\label{sec:interface}
\botheaderspace{}

Leveraging pre-computed multi-resolution embedding summarization~(\autoref{sec:method}), \tool{} tightly integrates three interface components~(\autoref{fig:crown}\figpart{A–C}).

\topheaderspace{}
\subsection{Map View}
\label{sec:interface:map}
\botheaderspace{}

The \mapview{}~(\autoref{fig:crown}\figpart{A}) is the primary view of \tool{}.
It provides a familiar map-like interface that allows users to pan and zoom to explore different embedding regions with varying sizes.
To help users easily investigate both the global structure and local neighborhoods of their embeddings, the \mapview{} integrates three layers of visualization.

\begin{figure}[tb]
  \includegraphics[width=\linewidth]{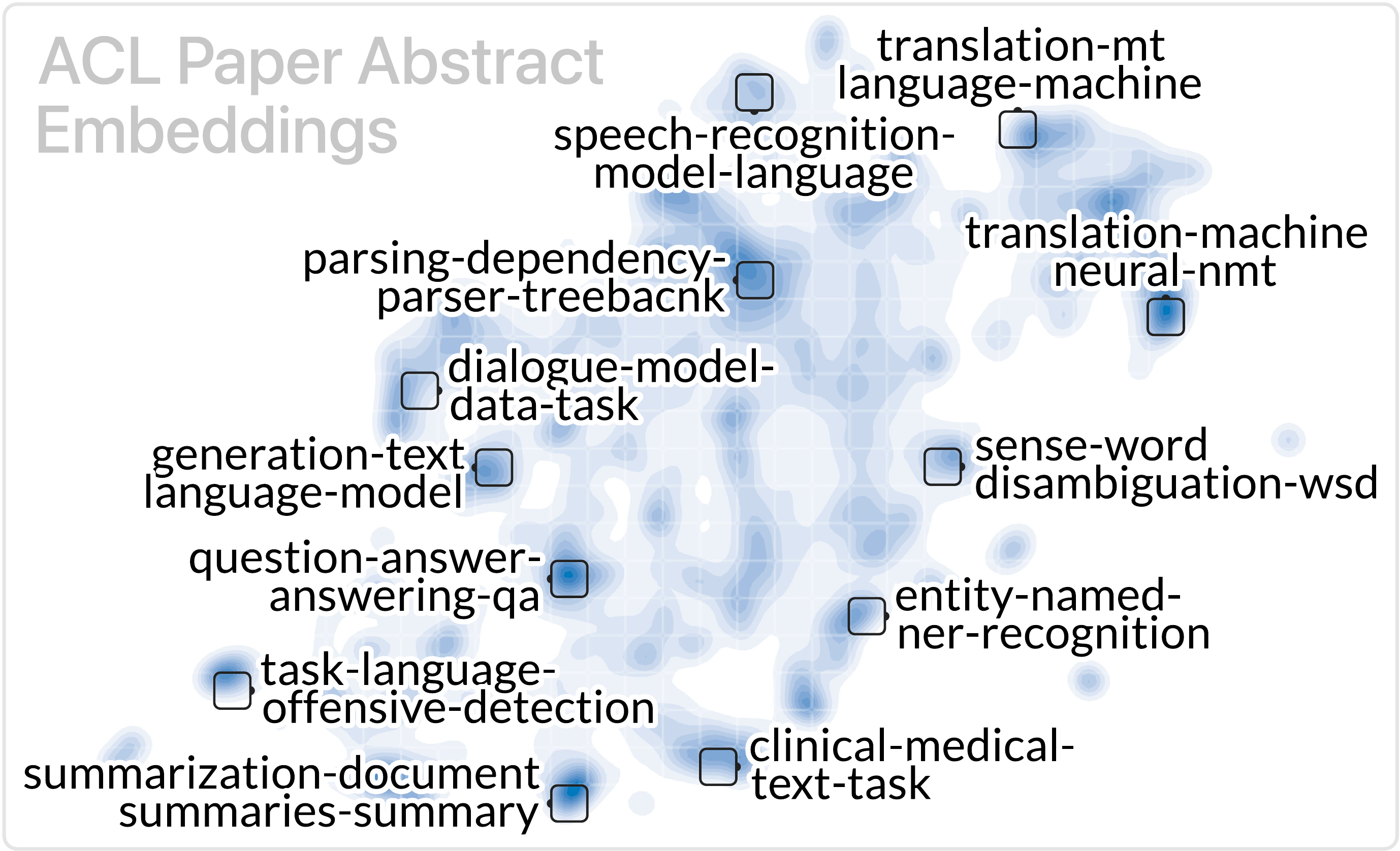}
  \centering
  \caption{
    The \mapview{} provides an embedding overview via a contour plot and auto-generated multi-resolution embedding labels placed around high-density areas.
  }
  \label{fig:contour}
\end{figure}

\paraspace{}
\paragraph{Distribution Contour.}
\label{sec:interface:map:contour}
To provide users with a quick overview of the global structure of their embeddings, we use Kernel Density Estimation (KDE)~\cite{rosenblattRemarksNonparametricEstimates1956} to estimate the distribution of 2D embedding points.
We use a standard multivariate Gaussian kernel with a Silverman bandwidth for the KDE model~\cite{silvermanDensityEstimationStatistics2018}.
Next, we compute the distribution likelihoods over a 200$\times$200 2D grid whose size is determined by the range of all embedding points.
Finally, we visualize the likelihoods over the grid as a contour plot~(\autoref{fig:contour}), highlighting the high-level density distribution of users' embeddings.
Researchers can adjust the grid density, and we tune it by balancing the computation time and the contour resolution.

\paraspace{}
\paragraph{Multi-resolution Labels.}
\label{sec:interface:map:label}
The \mapview{} helps users interpret embeddings across various levels of granularity by dynamically providing pre-computed contextual labels.
It overlays summaries generated via quadtree aggregation~(\autoref{sec:method}) onto the distribution contour and scatter plot.
Users can hover over to see the summary from a quadtree tile closest to the cursor.
Our tool adjusts the label's tile size based on the user's current zoom level.
For example, when a user zooms into a small region, the \mapview{} shows summaries computed at a lower level in the quadtree.
In addition to on-demand embedding summaries, this view also automatically labels high-density regions~(\autoref{fig:contour}) by showing summaries from quadtree tiles near the geometric centers of high-probability contour polygons.

\setlength{\columnsep}{10pt}
\setlength{\intextsep}{0pt}
\begin{wrapfigure}{R}{97pt}
  \vspace{-2pt}
  \centering
  \includegraphics[width=97pt]{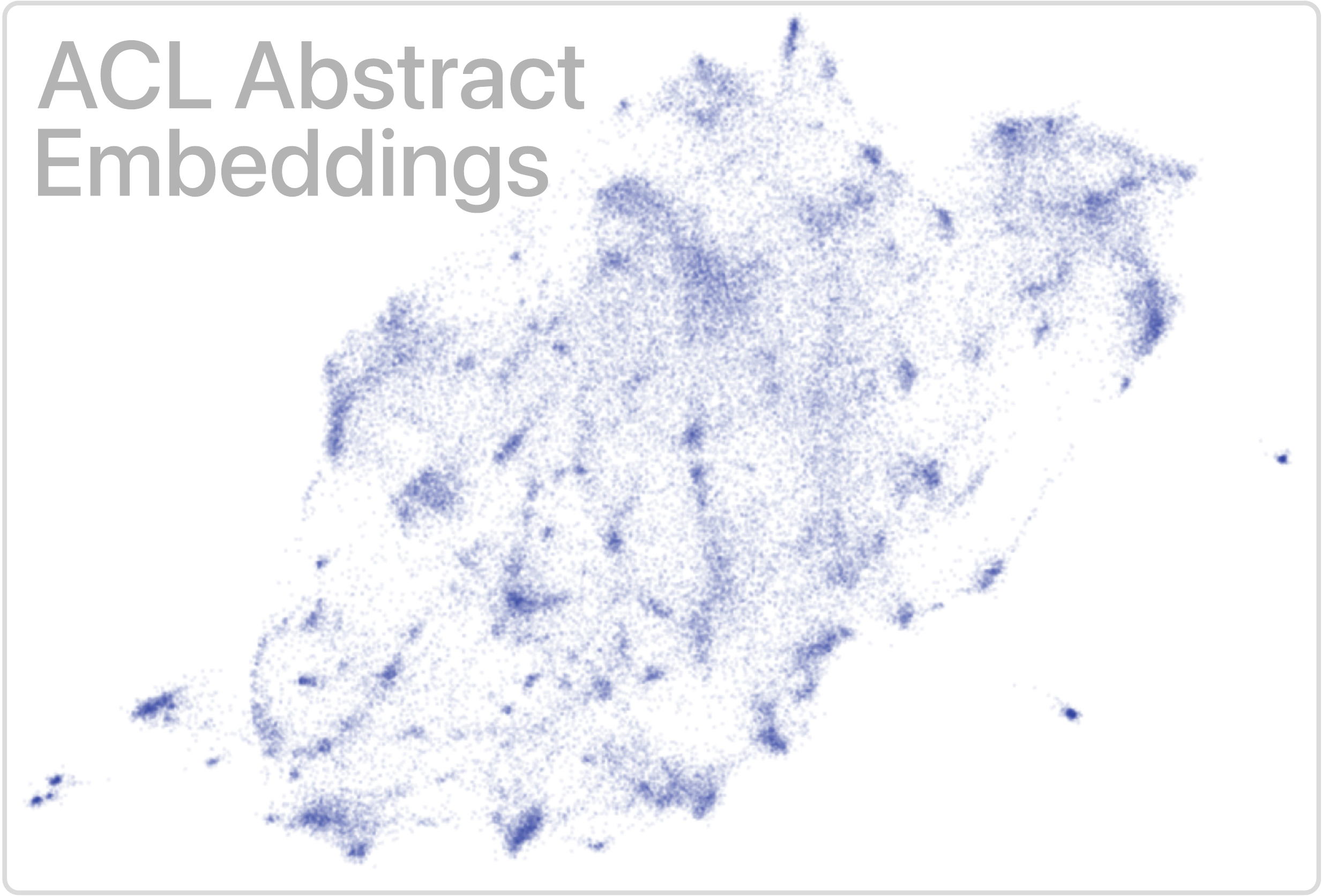}
  \vspace{-20pt}
  \vspace{0pt}
  \label{fig:score-tab}
\end{wrapfigure}
\paragraph{Scatter Plot.}
\label{sec:interface:map:scatter}
To help users pinpoint embeddings within their local neighborhoods, the \mapview{} visualizes all embedding points in a scatter plot with their 2D positions.
The corresponding scatter plot for \autoref{fig:contour} is shown on the right.
Users can specify the color of each embedding point to encode additional features, such as the class of embeddings.
Also, users can hover over an embedding point to reveal its original data, such as ACL paper abstracts~(\autoref{sec:senario:acl}).

\setlength{\abovecaptionskip}{3pt}
\setlength{\belowcaptionskip}{-12pt}
\setlength{\abovecaptionskip}{10pt}
\setlength{\belowcaptionskip}{0pt}

\topheaderspace{}
\setlength{\columnsep}{8pt}
\setlength{\intextsep}{0pt}
\begin{wrapfigure}{R}{105pt}
  \vspace{-2pt}
  \centering
  \includegraphics[width=105pt]{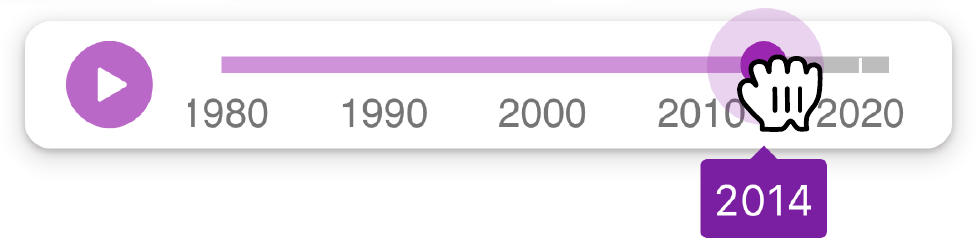}
  \vspace{-20pt}
  \vspace{0pt}
  \label{fig:score-tab}
\end{wrapfigure}
\subsection{Control Panel}
\label{sec:interface:control}
\begin{figure*}[tb]
  \includegraphics[width=\linewidth]{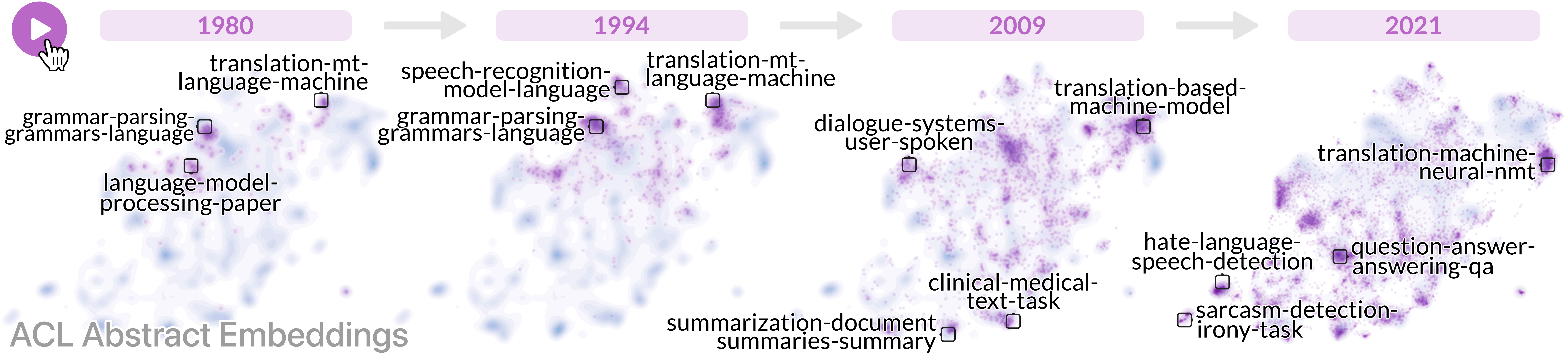}
  \centering
  \caption{
    \tool{} allows users to observe how embeddings change over time.
    For example, when exploring 63k ACL paper abstracts, clicking the play button \inlinefig{9}{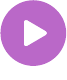} in the \controlpanel{} animates the visualizations to show embeddings of papers published in each year in \textcolor{purpleIV}{\textbf{purple}} and the distribution of all papers in \textcolor{indigoIII}{\textbf{blue}}.
    This animation highlights changes in ACL research topics over time, such as the decline in popularity of grammar and the rise of question-answering.
  }
  \label{fig:time-animation}
\end{figure*}

\botheaderspace{}
The \mapview{} shows all three visualization layers by default, and users can customize them to fit their needs by clicking buttons in the \controlpanel{}~(\autoref{fig:crown}\figpart{C}).
In addition, \tool{} allows users to compare multiple embedding groups in the same embedding space by superimposing them in the \mapview{}~\cite{gleicherConsiderationsVisualizingComparison2018}.
In the case of embeddings that include times, users can use a slider (shown on the right) in the \controlpanel{} to observe changes in the embeddings over time~(\autoref{fig:time-animation}).

\setlength{\abovecaptionskip}{3pt}
\setlength{\belowcaptionskip}{-12pt}
\setlength{\abovecaptionskip}{10pt}
\setlength{\belowcaptionskip}{0pt}

\topheaderspace{}
\subsection{Search Panel}
\label{sec:interface:search}
\botheaderspace{}

Searching and filtering can help users discover interesting embedding patterns and test their hypothesis regarding the embedding structure~\cite{carterActivationAtlas2019}.
In \tool{}, users can use the \serachpanel{}~(\autoref{fig:crown}\figpart{B}) to search text embeddings including specified words in the original data.
The panel shows a list of search results, and the \mapview{} highlights their corresponding embedding points.

\topheaderspace{}
\subsection{Scalable \& Open-source Implementation}
\label{sec:interface:implementation}
\botheaderspace{}

\tool{} is scalable to \textit{millions} of embedding points, providing a seamless user experience with zooming and animations, all within web browsers without backend servers.
To achieve this, we leverage modern web technologies, especially \linkhref{https://developer.mozilla.org/en-US/docs/Web/API/WebGL_API}{WebGL} to render embedding points with the \texttt{regl} API~\cite{lysenkoReglFunctionalWebGL2016}.
We also use \linkhref{https://developer.mozilla.org/en-US/docs/Web/API/Web_Workers_API/Using_web_workers}{Web Workers} and \linkhref{https://developer.mozilla.org/en-US/docs/Web/API/Streams_API}{Streams API} to enable the streaming of large embedding files in parallel with rendering.
To enable fast full-time search, we apply a contextual index scoring algorithm with \texttt{FlexSearch}~\cite{wilkerlingFlexSearchNextGenerationFull2019}.
We use \texttt{D3}~\cite{bostockDataDrivenDocuments2011} for other visualizations and \texttt{scikit-learn}~\cite{pedregosaScikitlearnMachineLearning2011} for KDE.
To ensure that our tool can be easily incorporated into users' current workflows~\cite{wangSuperNOVADesignStrategies2023}, we apply \texttt{NOVA}~\cite{wangNOVAPracticalMethod2022} to make \tool{} available within computational notebooks.
Users can also share their embedding maps with collaborators through unique URLs.
We provide detailed tutorials to help users use our tool with their embeddings.
We have open-sourced our implementation to support future research and development of embedding exploration tools. %

\topheaderspace{}
\section{Usage Scenarios}
\label{sec:senario}
\botheaderspace{}

We present two hypothetical scenarios, each with real embedding data, to demonstrate how \tool{} can help ML researchers and domain experts easily explore embeddings and gain a better understanding of ML model behaviors and dataset patterns.

\topheaderspace{}
\subsection{Exploring ACL Research Topic Trends}
\label{sec:senario:acl}
\botheaderspace{}

Helen, a science historian, is interested in exploring the evolution of computational linguistic and natural language processing (NLP) research since its inception.
She downloads the Bibtex files of all papers indexed in ACL Anthology~\cite{rohatgiACLAnthologyCorpus2022}. and extracts the paper title and abstract from 63k papers that have abstracts available.
Then, Helen applies MPNet, a state-of-the-art embedding model~\cite{songMpnetMaskedPermuted2020}, to transform the concatenation of each paper's title and abstract into a \texttt{768}-dimensional embedding vector.
She then trains a UMAP model to project extracted embeddings into a 2D space.
She tunes the UMAP's hyperparameter \texttt{n\_neighbors} to ensure projected points are spread out~\cite{coenenUnderstandingUMAP2019}.

Helen uses a Python function provided by \tool{} to generate three JSON files containing embedding summaries~(\autoref{sec:method}), the KDE distributions~(\autoref{sec:interface:map:contour}), and the original data in a streamable format~\cite{hoegerNewlineDelimitedJSON2014}.
Helen configures the function to use the dataset's \texttt{year} feature as the embedding's time---the function computes the KDE distribution of embeddings for each \texttt{year} slice.
She provides the files to \tool{} and sees a visualization of all ACL abstract embeddings~(\autoref{fig:contour}\figpart{A}).

\paraspace{}
\paragraph{Embedding Exploration.}
In the \mapview{}, Helen explores embeddings with zoom and pan.
She also uses the \serachpanel{} to find papers with specific keywords, such as ``dialogue'', and Helen is pleased to see all related papers are grouped in a cluster~(\autoref{fig:crown}\figpart{B}).
With the help of multi-resolution embedding summaries, Helen quickly gains an understanding of the structure of her embedding space.
For example, she finds that the top right cluster features translation papers while the lower clusters feature summarization and medical NLP.

\paraspace{}
\paragraph*{Embedding Evolution}
To examine how ACL research topics change over time, Helen clicks the play button clicking the play button \inlinefig{9}{icon-play-solid-paper} in the \controlpanel{} to animate the visualizations.
The \mapview{} shows embeddings of papers published in each year from 1980 to 2022 in \textcolor{purpleIV}{\textbf{purple}}, while the distribution of all papers is shown as a \textcolor{indigoIII}{\textbf{blue background}}~(\autoref{fig:time-animation}).
As Helen observes the animation, she identifies several interesting trends.
For example, she observes a decline in the popularity of grammar research, while question-answering has become increasingly popular.
She also notes the emergence of some small clusters in recent years, featuring relatively new topics, such as sarcasm, humor, and hate speech.
Satisfied with the findings using \tool{}, Helen decides to write an essay on the trend of NLP research over four decades.

\topheaderspace{}
\subsection{Investigating Text-to-Image Model Usage}
\label{sec:senario:diffusiondb}
\botheaderspace{}

Bob, an ML researcher, works on improving text-to-image generative models.
Recent advancements in diffusion models, such as Stable Diffusion~\cite{rombachHighresolutionImageSynthesis2022}, have attracted an increasing number of users to generate photorealistic images by writing text prompts.
To gain an understanding of these models' behaviors and identify potential weaknesses for improvement, Bob decides to study how users use these models in the wild by analyzing DiffusionDB, a dataset containing 14 million images generated by Stable Diffusion with 1.8 million unique text prompts~\cite{wangDiffusionDBLargescalePrompt2022}.

\begin{figure}[tb]
  \includegraphics[width=\linewidth]{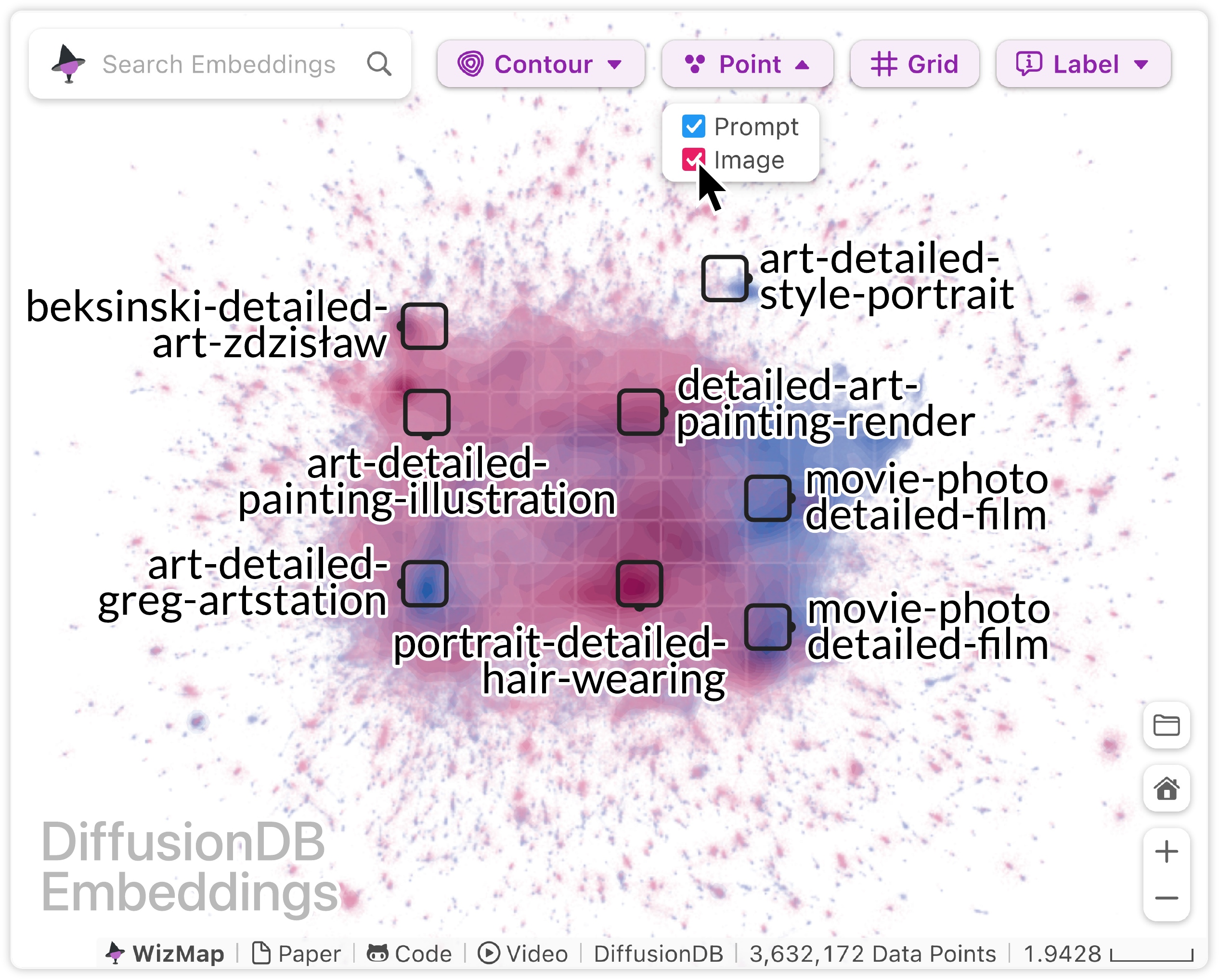}
  \caption{
    \tool{} enables users to compare multiple embeddings by visualization superposition.
    For instance, comparing the CLIP embeddings of \textbf{1.8 million} Stable Diffusion \textcolor{indigoIV}{\textbf{prompts}} and \textbf{1.8 million} generated \textcolor{pinkIII}{\textbf{images}} reveals key differences between two distributions.
  }
  \label{fig:diffusiondb}
\end{figure}

Bob's analysis goal is to study the relationship between the text prompts and their generated images.
Thus, he chooses to use CLIP~\cite{radfordLearningTransferableVisual2021} to encode both prompts and images into a \texttt{768}-dimensional multimodal embedding before projecting them to a 2D space with UMAP.
He uses prompts to generate embedding summaries for the CLIP space.
After creating all JSON files, \tool{} visualizes all 3.6 million embeddings~(\autoref{fig:diffusiondb}).

\paraspace{}
\paragraph{Embedding Exploration.}
Bob begins his exploration by hiding image embeddings and scatter plots, focusing on the global structure of embeddings with the contour plot and embedding summaries.
He discovers two dominant prompt categories: art-related prompts and photography-related prompts.
Two categories appear far from each other, which is not surprising as they are expected to have distinct semantic representations.
Bob also notices two smaller clusters within the photography region, prompting him to zoom in and turn on the scatter plot to further investigate these local regions~(\autoref{fig:transition}).
After hovering over a few points, he realizes one cluster is mostly about non-human objects while the other is about celebrities.

\paraspace{}
\paragraph{Embedding Comparison.}
To investigate the relationship between text prompts and their generated images, Bob clicks a button in the \controlpanel{} to superimpose the contour and scatter plot of image embeddings \textcolor{pinkIII}{\textbf{in red}} onto the text embedding visualizations \textcolor{indigoIV}{\textbf{in blue}}~(\autoref{fig:diffusiondb}).
Bob quickly identifies areas where two distributions overlap and differ.
He notes that the ``movie'' cluster in the text embeddings has a lower density in the image embeddings, whereas a high-density ``art portrait'' cluster emerges in image embeddings.
Bob hypothesizes that Stable Diffusion may have limited capability to generate photorealistic human faces~\cite{borjiGeneratedFacesWild2022}.
After exploring embedding with \tool{}, Bob is pleased with his findings, and he will apply his insights to improve the curation of his training data.

\setlength{\abovecaptionskip}{2pt}
\setlength{\belowcaptionskip}{-12pt}
\setlength{\abovecaptionskip}{10pt}
\setlength{\belowcaptionskip}{0pt} %
\topheaderspace{}
\section{Future Work and Conclusion}
\botheaderspace{}
\vspace{-2pt}

\tool{} integrates a novel quadtree-based embedding summarization technique that enables users to easily explore and interpret large embeddings across different levels of granularity.
Our usage scenarios showcase our tool's potential for providing ML researchers and domain experts with a holistic view of their embeddings.
Reflecting on our design and development of \tool{}, we acknowledge its limitations and distill future research directions that could further assist users in interpreting and applying embeddings for downstream tasks.

\begin{itemize}[topsep=2pt, itemsep=0mm, parsep=2pt, leftmargin=9pt]
  \item \textbf{User evaluation.}
  To investigate the usefulness of flexible transitioning across various levels of abstraction during embedding exploration, future researchers can use \tool{} as a research instrument to conduct observational user studies with ML researchers and domain experts.

  \item \textbf{Automated insights.}
  Our tool provides automatic and multi-scale visual contexts to guide users in their exploration.
  While our quadtree-based approach is effective and scalable, it is sensitive to tile size selection.
  Researchers can explore more robust methods for embedding summarization and automated data insights, such as clustering-based approaches~\cite{lawCharacterizingAutomatedData2020}.

  \item \textbf{Enhanced comparison.}
  \tool{} helps users compare embedding groups through contour superposition.
  However, for local comparisons, other techniques such as juxtaposition and explicit encoding may be more effective~\cite{gleicherConsiderationsVisualizingComparison2018}.
  Future researchers can design visualization tools that integrate these techniques.

\end{itemize}

\section{Broader Impact}
We designed and develop \tool{} with good intentions---to help ML researchers and domain experts easily explore and interpret large embeddings.
However, bad actors could exploit insights gained from using \tool{} for malevolent purposes.
For example, research has shown that ML embeddings contain societal biases~\cite{bolukbasiManComputerProgrammer2016}.
Therefore, bad actors could manipulate and sabotage ML predictions by injecting inputs whose embeddings are known to associate with gender and racial biases.
The potential harms of biased embeddings warrant further study.

\section*{Acknowledgements}

We thank our anonymous reviewers for their insightful comments.
This work was supported in part by a J.P. Morgan PhD Fellowship, Apple Scholars in AI/ML PhD fellowship, DARPA GARD, gifts from Cisco, Bosch, and NVIDIA.
Use, duplication, or disclosure is subject to the restrictions as stated in Agreement number HR00112030001 between the Government and the Performer.

\bibliographystyle{acl_natbib}
\bibliography{23-wizmap}

\end{document}